\documentclass[journal]{IEEEtran}
\usepackage{blindtext}
\usepackage{graphicx, subcaption}
\usepackage[numbers]{natbib}
\usepackage{xcolor}
\usepackage{amsmath}
\usepackage{url}
\ifCLASSINFOpdf
\else
\fi

\usepackage{romannum}

\begin{document}
\pagenumbering{arabic} 
%
\title{LS-Net: Fast Single-Shot Line-Segment Detector}
%
%
%

\author{\IEEEauthorblockN{Van Nhan Nguyen\IEEEauthorrefmark{1}\IEEEauthorrefmark{2},
Robert Jenssen\IEEEauthorrefmark{1}, and
Davide Roverso\IEEEauthorrefmark{2}}

\IEEEauthorblockA{\IEEEauthorrefmark{1}The UiT Machine Learning Group, UiT The Arctic University
of Norway, 9019 Troms{\o}, Norway}

\IEEEauthorblockA{\IEEEauthorrefmark{2}Analytics Department, eSmart Systems, 1783 Halden, Norway}
}

\maketitle

\begin{abstract}
In low-altitude Unmanned Aerial Vehicle (UAV) flights, power lines are considered as one of the most threatening hazards and one of the most difficult obstacles to avoid. In recent years, many vision-based techniques have been proposed to detect power lines to facilitate self-driving UAVs and automatic obstacle avoidance. However, most of the proposed methods are typically based on a common three-step approach: (i) edge detection, (ii) the Hough transform, and (iii) spurious line elimination based on power line constrains. These approaches not only are slow and inaccurate but also require a huge amount of effort in post-processing to distinguish between power lines and spurious lines. In this paper, we introduce LS-Net, a fast single-shot line-segment detector, and apply it to power line detection. The LS-Net is by design fully convolutional and consists of three modules: (i) a fully convolutional feature extractor, (ii) a classifier, and (iii) a line segment regressor. Due to the unavailability of large datasets with annotations of power lines, we render synthetic images of power lines using the Physically Based Rendering (PBR) approach and propose a series of effective data augmentation techniques to generate more training data. With a customized version of the VGG-16 network as the backbone, the proposed approach outperforms existing state-of-the-art approaches. In addition, the LS-Net can detect power lines in near real-time (20.4 FPS). This suggests that our proposed approach has a promising role in automatic obstacle avoidance and as a valuable component of self-driving UAVs, especially for automatic autonomous power line inspection.
\end{abstract}



%
\IEEEpeerreviewmaketitle

\section{Introduction}
Obstacle detection and avoidance are key to ensure low altitude fight safety. Due to their extremely small size, power lines are considered as one of the most threatening hazards and one of the most difficult obstacles for Unmanned Aerial Vehicles (UAVs) to avoid \cite{shan2017multiple}.

In automatic autonomous vision-based power line inspection, power line detection is crucial. Not only for ensuring flight safety, and for vision-based navigation of UAVs, but also for inspection to identify faults on power lines (e.g., corroded and damaged power lines) and surrounding objects, such as vegetation encroachment \cite{nguyen2018automatic}. 

In recent years, many techniques have been proposed to detect power lines automatically. However, most of the proposed methods are typically based on a common three-step approach: First, an edge detector such as Canny \cite{canny1986computational} is applied to produce edge maps. Then, the Hough transform \cite{duda1972use}, the Radon transform, or a line tracing algorithm, are utilized to detect straight lines from the edge maps. Finally, power line constraints, such as parallel lines, are applied to eliminate spurious lines and detect the power lines. These approaches not only are slow and inaccurate but also require a considerable amount of effort in post-processing to distinguish between power lines and spurious lines. 

With the aim of facilitating real-time and accurate power line detection for UAV vision-based navigation and inspection, we propose in this paper LS-Net, a fast single-shot line-segment detector, and apply it to power line detection. 

The work presented in this paper is part of an ongoing effort involving the exploitation of recent advances in Deep Learning (DL) and UAV technologies for facilitating automatic autonomous vision-based inspection of power lines. In our previous work \cite{nguyen2018automatic}, we first proposed a novel automatic autonomous vision-based power line inspection concept that uses  UAV inspection as the main inspection method, optical images as the primary data source,  and deep learning as the backbone of data analysis. We then identified six main challenges of DL vision-based UAV inspection: the lack of training data; class imbalance; the detection of small power components and faults; the detection of power lines in cluttered backgrounds; the detection of previously unseen components and faults; and the lack of metrics for evaluating inspection performance. 

To move forward, we proposed approaches to address the first three challenges and built a basic automatic vision-based inspection system with two custom-built UAVs and five DL-based models for data analysis and inspection \cite{nguyen2019intelligent}. 

In this paper, we take this further by addressing the fourth challenge of DL vision-based UAV inspection, which is to detect power lines in cluttered backgrounds, with our proposed LS-Net. The LS-Net is a feed-forward, fully Convolutional Neural Network (CNN) \cite{springenberg2015striving}, and consists of three modules: (i) a fully convolutional feature extractor, (ii) a classifier, and (iii) a line segment regressor. Due to the unavailability of large datasets with annotations of power lines, we render synthetic images of power lines using the Physically Based Rendering (PBR) approach \cite{pharr2016physically} and propose a series of effective data augmentation techniques to generate more training data. With a customized version of the VGG-16 network \cite{simonyan2014very} as the backbone, the proposed LS-Net outperforms existing state-of-the-art DL-based power line detection approaches and shows the potential to facilitate real-time power line detection for obstacle avoidance in low-altitude UAV flights.

The contribution of this paper is fourfold. Firstly, we propose a novel single-shot line segment detector, called LS-Net. The proposed LS-Net can be trained end-to-end via a weighted multi-task loss function, which is a combination of Focal loss \cite{lin2017focal} for addressing the class imbalance in classification and Wing loss  \cite{feng2018wing} for restoring the balance between the influence of errors of different sizes in multiple points regression. Secondly, we resolve the issues of single-shot detectors, which typically employ a traditional one-grid approach, when applied to line segment detection by proposing a four-grid approach. To the authors knowledge, such an approach is new in the single-shot approaches based on CNNs. Thirdly, we address the lack of training data by using synthetic data rendered by the PBR approach and applying a series of effective data augmentation techniques to generate more training data. Finally, this work is in our opinion paving the way for fully automatic autonomous vision-based power line inspection, in which high-speed  UAVs equipped with sensors, cameras, a DL vision-based UAV navigator, and a DL-based model for data analysis, can automatically navigate along power lines to collect data for offline inspections and perform online inspections to identify potential faults quickly.

The remainder of the paper is structured as follows: Section \ref{sec:relatedwork} presents background knowledge and relevant related work in CNN-based image classification models, CNN-based object detection models, and common approaches to power line detection. Then we describe our proposed LS-Net, a fast single-shot line segment detector in Section \ref{sec:lsnet}. Next, in section \ref{sec:experiments}, we present in detail our experimental results and ablation studies. Further, in Section \ref{sec:discussion}, we discuss the potential of our proposed LS-Net in UAV navigation and inspection as well as in detecting other linear structures, such as railway tracks, unburied onshore pipelines, and roads. Finally, in Section \ref{sec:conclusion}, we conclude the paper with a summary and an outlook for the future of the field.

\section{Background and Related Work}
\label{sec:relatedwork}
\subsection{Convolutional Neural Networks}
In the past few years, Convolutional Neural Networks (CNNs) \cite{krizhevsky2012imagenet}, which are special neural networks designed to take advantage of the 2D structure of image data, have been advancing the state-of-the-art for both high-level tasks, such as image classification, object detection, image segmentation, and low-level vision tasks, for instance edge detection. In this section, we summarize some of the most well-known CNN architectures for those tasks and describe a selection of methods and techniques that will be used in the LS-Net.
\subsubsection{High-level vision tasks}
Since the success of Krizhevsky et al. \cite{krizhevsky2012imagenet} with an 8-layer CNN (5 convolutional layers + 3 fully-connected layers) in the 2012 ImageNet challenge, CNNs have become a commodity in the computer vision field. In the last few years, many attempts have been made to improve the original architecture of  Krizhevsky et al. by, for example, utilizing smaller receptive window size and by increasing the depth of the network.

One of the most recognized such attempts is the VGGNet, which is a CNN architecture that secured the first and the second places in the localization and classification tasks, respectively, in the 2014 ImageNet challenge \cite{simonyan2014very}. The key innovation of the VGGNet is the combination of small filters ($3\times 3$ filters) and deep networks (16-19 layers). The authors argued that a stack of three $3\times 3$ convolutional layers has the same effective receptive field as
one $7\times 7$ convolutional layer, but is deeper, has more non-linearities and fewer parameters.

With the increasing complexity of image classification problems, deeper CNNs are typically required. However, deep CNNs constructed simply by stacking up many layers are very difficult to train due to the notorious problem of vanishing/exploding gradients. To ease the training of deep CNNs, Residual Networks (ResNets) were proposed \cite{he2016deep}. ResNets add ``shortcut'' connections to the standard CNN layers to allow the gradient signal to travel back directly from later layers to early layers. The ``shortcut'' connections allowed the authors of ResNets to successfully train very deep CNNs with 50, 101, and even 152 layers.

Inspired by the success of CNNs in image classification, Faster R-CNN (Region-based Convolutional Neural Network) was proposed to solve a more challenging task of object detection \cite{ren2015faster}. Faster R-CNN is a single, unified network that performs object detection via two main steps: region proposal and region classification. First, a base network (e.g., ResNet \cite{he2016deep}) is utilized to extract features from images. Next, the extracted features are fed into a Region Proposal Network (RPN) to find proposals. Then,  a CNN-based classifier is applied on top of the extracted feature maps to classify the proposals and refine their bounding boxes. Finally, post-processing is used to refine the bounding boxes further and eliminate duplicate detections. Faster R-CNN is very accurate; however, it is quite slow. 

R-FCN (Region-based Fully Convolutional Network)  \cite{dai2016r} is an accurate and efficient object detection framework proposed to address existing issues of region-based detectors such as Fast R-CNN \cite{girshick2015fast} and Faster R-CNN \cite{ren2015faster}. Instead of applying a costly per-region sub-network hundreds of times, R-FCN adopts a fully convolutional architecture with almost all computations shared across the entire image. To address the dilemma between translation-invariance in image classification and translation-variance in object detection, R-FCN proposes novel position-sensitive score maps which allow fully convolutional networks to effectively and efficiently perform both classification and detection in a single evaluation. With those novel improvements, R-FCN can run at 2.5-20 times faster and achieve higher accuracy than the Faster R-CNN counterpart. RPN based approaches are accurate; however, they are typically slow due to their complex multi-stage pipelines \cite{redmon2016you}. With the aim of facilitating real-time object detection, many single-shot object detectors, which take only one single-shot to detect multiple objects in the image, have been proposed. The two most well-known single-shot object detectors are YOLO \cite{redmon2016you} and SSD \cite{liu2016ssd}.

YOLO (You Only Look Once) is a real-time object detection framework that directly predicts bounding boxes and class probabilities with a single network in a single evaluation \cite{redmon2016you}. To achieve this, YOLO unifies region proposal and region classification into a single neural network and, according to the authors, ``frames object detection as a regression problem to spatially separated bounding boxes and associated class probabilities''. YOLO divides the input image into a $S\times S$ grid. Each grid cell predicts $B$ bounding boxes, confidence scores for those boxes, and $C$ conditional class probabilities. With a unified architecture, YOLO is extremely fast; it processes images in real-time. However, YOLO is not state-of-the-art in terms of accuracy.

SSD (Single-Shot MultiBox Detector) improves YOLO by adding a series of modifications: (i) a small convolutional filter is utilized to predict object classes and offsets in bounding box locations;  separate predictors (filters) are employed for predicting objects at different aspect ratios; predictions are performed at multiple feature maps from the later stages of a network to enable detection at multiple scales \cite{liu2016ssd}. These modifications make SSD both faster and more accurate than the YOLO counterpart.

Lin et at. identified class imbalance during training as the main obstacle preventing one-stage detectors (e.g., SSD and YOLO) from achieving the state-of-the-art accuracy and proposed to address that by introducing a novel loss function named Focal Loss (FL) \cite{lin2017focal}. FL dynamically scales the standard cross-entropy loss with a scaling factor that decays to zero as confidence in the correct class increases. By doing that, FC automatically reduces the weights of easy examples during training and allows the model to focus on hard examples. 

Feng et al. observed that in multiple points localization problems, such as facial landmark localization, more attention should be paid to the samples with small or medium range errors \cite{feng2018wing}. To achieve this target, the authors proposed a new loss function, namely, Wing loss. With the aim of restoring the balance between the influence of errors of different sizes, Wing loss was designed to behave as a log function with an offset for small errors and as L1 for large errors. According to the authors, Wing loss is appropriate for dealing with relatively small localization errors.

Ioffe et al. observed that the change in the distributions of layers' inputs during the training of deep neural networks poses a serious problem because the layers need to adapt to the new distribution continuously \cite{ioffe2015batch}; this phenomenon was referred to as \textit{internal covariate shift}. To address this problem, the authors proposed a new mechanism, called \textit{Batch Normalization (BN)}, which fixes the means and variances of layer inputs by normalizing each activation independently along the batch dimension. 

Although the normalization along the batch dimension allows BN to reduce internal covariate shift and accelerate the training of deep neural nets, it causes many distinct drawbacks. For example, for BN to work properly, it is required to have a sufficiently large batch size (e.g., 32 per worker), which is typically not possible with very deep CNNs and high-resolution images due to GPU memory limitations  \cite{GroupNorm2018}. With the aim of eliminating the dependence on batch sizes and avoiding batch statistics computation,  Wu et al. proposed \textit{Group Normalization (GN)} as a simple alternative to BN \cite{GroupNorm2018}. The key innovation of GN is that it divides channels into groups and normalizes the features within each group. 

\subsubsection{Low-level vision tasks}
CNNs have been successfully applied to low-level vision tasks such as edge detection. For example, Xie et al. proposed a method, named holistically-nested edge detection (HED),  for predicting edges in an image-to-image fashion \cite{xie2015holistically}. The method leverages fully convolutional neural networks and deeply-supervised nets by attaching a side output layer to the last convolutional layer in each stage of the VGGNet and utilizing both weighted-fusion supervision and deep supervision in training. Liu et al. improved the HED method by utilizing richer features from all convolutional layers in the VGGNet \cite{liu2017richer}. In addition, a novel loss function was proposed to treat training examples properly, and a multi-scale hierarchy was employed to enhance edges.

CNNs have also been successfully applied to semantic line detection. For example, Lee et at. proposed a semantic line detector (SLNet) based on the VGG16 network \cite{lee2017semantic}. First, multi-scale feature maps are extracted from an input image using convolution and max-pooling layers. Then, a line pooling layer is developed to extract a feature vector for each line candidate. Finally, the feature vectors are fed into parallel classification and regression layers to decide whether the lines are semantic or not and to refine their location. 

\subsection{Common approaches to power line detection}
\label{sec:commonapproachestopld}
\subsubsection{Line-based methods}
A straight-forward approach to power line detection is to treat the power line as a straight line and apply line detection algorithms directly. For example, Li et al. utilized the Hough transform to detect straight lines from Pulse Coupled Neural Network filtered images and employed K-means clustering to discriminate power lines from other mistakable linear objects \cite{li2008knowledge,li2010towards}. 

Although this approach is effective and easy to implement, its strong assumptions on the characteristics of power lines, including, (i) a power line has uniform brightness, (ii) a power line approximates a straight line, and (iii) power lines are approximately parallel to each other, make it a less practical approach. Due to the strong assumptions, line-based methods often mistakenly detect linear objects, such as metallic fence lines \cite{li2008knowledge}, as power lines and misdetected power lines that appear as arc curves due to the influence of gravity \cite{shan2017multiple}. 

\subsubsection{Piece-wise line segment-based methods}
With the aim of detecting both straight power lines and curvy ones, some researchers have proposed to segment a power line into piece-wise line segments so that they can be approximated by straight lines \cite{yan2007automatic,song2014power}. For example, Yan et al. utilized the Radon transform to extract line segments of a power line, then employed a grouping method and the Kalman filter to link each line segment and connect the linked line segments into a complete line \cite{yan2007automatic}. Song et at. applied matched filter and first-order derivative of Gaussian to detect line segments, then used a graph cut model based on graph theory to group the detected line segments into whole power lines \cite{song2014power}. 

Similar to line-based methods, piece-wise line segment-based methods also often mistakenly detect linear objects with similar line features in the background, such as metallic fence lines and building edges, as power lines \cite{song2014power}. 

\subsubsection{Auxiliaries assisted methods}
To address the existing problems of line-based and piece-wise line segment-based methods, much effort has been made towards utilizing correlation information and context features provided by auxiliaries. For example, Zhang et al. proposed to use the spatial correlation between the pylon and the power line to improve transmission line detection performance \cite{zhang2014pylon}. The proposed method outperforms line-based and piece-wise line segment-based methods; however, the performance drops significantly when the pylon is absent or occluded.

To eliminate the need for manually selecting auxiliaries and defining spatial relationships between auxiliaries and power lines, Shan et at. proposed an optimization-based approach for automatic auxiliaries selection and contexts acquisition \cite{shan2017multiple}. The proposed approach surpasses traditional methods that use manually assigned auxiliaries both in terms of detection accuracy and false alarm probability; however, it is quite slow due to the sliding window-based object extraction and the context representation between the auxiliaries and the hypotheses. 

To further improve auxiliaries assisted power line detection accuracy and speed, Pan et al. proposed a metric for measuring the usefulness of an auxiliary in assisting power line detection, named spacial context disparity, based on two factors: spatial context peakedness and spatial context difference and applied it for automatic selection of optimal auxiliaries \cite{pan2017leveraging}.  According to the authors, the proposed method is robust and can achieve satisfactory performance for power line detection. 

\subsubsection{DL-based methods}
One of the earliest attempts to use deep learning for power line detection was the work of Jayavardhana et al. \cite{gubbi2017new}. The authors proposed a CNN-based classifier that uses Histogram of Gradient (HoG) features as the input and applied it in a sliding window fashion to classify patches of size $32\times 32$ into two classes: ``Line present'' and ``No line present''. The authors also finetuned the GoogleNet on patches of original images for the same task. According to the authors, the proposed CNN-based classifier achieves an F-score of 84.6\% and outperforms the GoogleNet, which achieves an F-score of 81\%.

Ratnesh et al. \cite{madaan2017wire} treated wire detection as a semantic segmentation task and performed a grid search over a finite space of CNN architectures to find an optimal model for the task based on dilated convolutional networks \cite{yu2015multi}. The model was trained on synthetic images of wires generated by a ray-tracing engine and finetuned on real images of wires from the USF dataset \cite{kasturi2002wire}. According to the authors, the proposed model outperforms previous work that uses traditional computer vision and various CNN-based baselines such as FCNs, SegNet, and E-Net; the model achieves an Average Precision (AP) score of 0.73 on the USF dataset and runs at more than 3Hz on the NVIDIA Jetson TX2 with input resolution of $480\times 640$. 

Although treating wire detection as a semantic segmentation task has been proved to be a powerful approach for detecting wires \cite{madaan2017wire}, its requirement of pixel-level annotated ground-truth data makes it less practical than traditional computer vision approaches. Sang et al. proposed to use weekly supervised learning with CNNs for localizing power lines in pixel-level precision by only using image-level class information \cite{lee2017weakly}. First, a classifier adapted from the VGG19 is applied to classify sub-regions ($128\times 128$) from an input image ($512\times 512$) by using a sliding window approach. Then, feature maps of intermediate convolutional layers of sub-regions that are classified as ``sub-region with power lines'' are combined to visualize the location of the power lines. Although the localization accuracy of the proposed approach is still far from an applicable level of industrial fields, it can be applied, according to the authors, to generate ground-truth data in pixel-level roughly.

Yan et al. proposed a power detection pipeline based on pyramidal patch classification in \cite{li2018power}. First, input images are hierarchically partitioned into patches. Next, a CNN classifier is trained to classify the patches into two classes: patches with power lines and patches without power lines. Then, the classified patches are used as inputs for edge feature extraction using steerable filters and line segment detection using the Progressive Probabilistic Hough Transform (PPHT). Finally, the detected line segments are connected using a power line segments correlation module to form complete power lines. The authors concluded that the proposed approach significantly improves the detection rate of the power line detection and largely decreases the false alarm rate.

\section{The Line Segment Detector (LS-Net)}
\label{sec:lsnet}
\subsection{Data Generation}
\subsubsection{Synthetic Data Generation}


Due to the unavailability of large datasets with annotations of power lines, we collaborate with Nordic Media Lab (NMLab)\footnote{\url{http://nmlab.no/}} to render synthetic images of power lines using the Physically Based Rendering (PBR) approach \cite{pharr2016physically}. First, we model Aluminium Conductor Steel-Reinforced (ACSR) cables, which are typically composed of one steel center strand and concentric layers of high-purity aluminum outer strands, using the Autodesk 3DS Max program. To increase the realistic appearance of the cables, we utilize the bevel and the twist modifiers together with the metal brushed steel texture. Then, we randomly superimpose the cables on 71 8K High Dynamic Range Images (HDRIs) collected from the internet\footnote{https://hdrihaven.com/}. Next, to further increase the realistic appearance of the cables, we employ cube mapping to capture the reflection and the lighting data from the HDRIs and apply them to the cables. Finally, we apply a series of effective variations, with respect to the camera angle, the camera distance, out-of-focus blur, cable colors, the number of cables, and the distance between cables, to render more synthetic images.

\subsubsection{Data Augmentation}


Inspired by the success of data augmentation for improving the performance of CNNs in \cite{wong2016understanding}, \cite{howard2013some}, and \cite{simard2003best}, we propose a series of effective data augmentation techniques to generate more training data by applying transformations in the data-space. These are all implemented using the scikit-image \cite{scikit-image} and the OpenCV libraries \cite{opencv_library}. 

\begin{figure*}[t]
    \centering
    \includegraphics[width=1.0\textwidth]{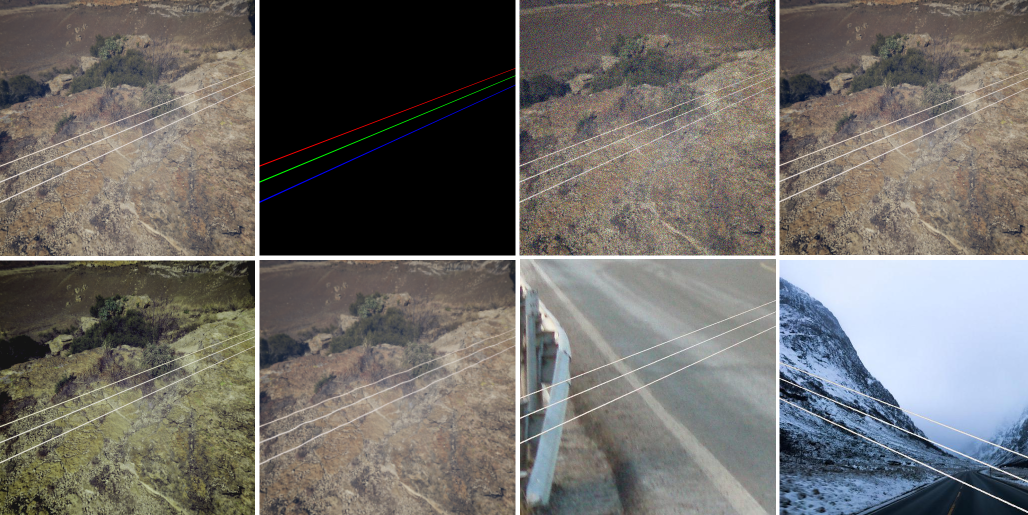}
    \caption{Sample augmented images (from left to right): original image;  pixel-level annotation, image with Gaussian-distributed additive noise; Gaussian blurred image; color manipulated image; elastic transformed image, image with new background, cropped and flipped image with new background.}
    \label{fig:pipeline}
\end{figure*}

The first technique replaces the background of the generated synthetic images with real background images to increase the diversity of the dataset and to account for various types of background variations during the inspection (e.g., different seasons, weather conditions, and lighting conditions).

The second technique adds Gaussian-distributed additive noise to account for noisy image acquisition (e.g., sensor noise caused by poor illumination and/or high temperature, and/or transmission) \cite{boyat2015review}. The augmented image $f(i,j)$ is the sum of the true image $s(i,j)$ and the noise $n(i,j)$:
\begin{equation}
    f(i,j) = s(i,j) + n(i,j).
\end{equation}
The noise term, $n(i, j)$, follows a Gaussian random distribution:
\begin{equation}
    p_{G}(z)={\frac  {1}{\sigma {\sqrt  {2\pi }}}}e^{{-{\frac  {(z-\mu )^{2}}{2\sigma ^{2}}}}},
\end{equation}
where $z$ represents the gray level,  $\mu$  is the mean value, and $\sigma$  is the standard deviation.

To account for possible out-of-focus, Gaussian blur is employed by convolving the image with a two-dimensional Gaussian function:
\begin{equation}
\label{equation:2dgaussian}
    G(x,y)={\frac {1}{2\pi \sigma ^{2}}}e^{-{\frac {x^{2}+y^{2}}{2\sigma ^{2}}}},
\end{equation}
where $x$ and $y$ are distances from the origin in the horizontal axis and the vertical axis respectively, and $\sigma$ is the standard deviation \cite{shapiro2001computer}.

To introduce invariance to changes in lighting and to capture minor color variations, especially in power lines, a series of color manipulations including random brightness, random saturation, random contrast, and random hue are utilized. In addition, to further extend color invariance, we randomly remove colors from RGB images by first converting them to grayscale and then converting the grayscale images back to RGB.

With the aim of training models that can detect not only perfectly straight line segments but also curvy ones, elastic deformations \cite{simard2003best} are employed. First, two random displacement fields for the $x$-axis ($\Delta x$) and $y$-axis ($\Delta y$) are generated as follows:
\begin{gather}
    \Delta x (x,y) = rand(-1,+1), \\
    \Delta y (x,y) = rand(-1,+1),
\end{gather}
where $rand(-1,1)$ is a random number between $-1$ and $+1$, generated with a uniform distribution. Next, the fields $\Delta x$ and $\Delta y$ are convolved with  a two-dimensional Gaussian function similar as shown in Eq. (\ref{equation:2dgaussian}) to form elastic deformation fields. Then, the elastic deformation fields are scaled by factor $\alpha$ that controls the intensity of the deformation. Finally, the fields $\Delta x$ and $\Delta y$ are applied to images.

To account for various camera distances and viewing angles, zoom and rotation operators are employed \cite{wolberg1990digital}. The zoom operator is applied by randomly cropping images and scaling them to their original size. The rotation operator is employed by multiplying images with a rotation matrix $R$: 
\[
R={\begin{bmatrix}\cos \theta &-\sin \theta \\\sin \theta &\cos \theta \\\end{bmatrix}},
\]
where $\theta$ is the rotation angle.
The final technique flips the images horizontally and vertically.

\subsection{LS-Net Architecture}
\begin{figure}[h!]
    \centering
    \includegraphics[width=0.40\textwidth]{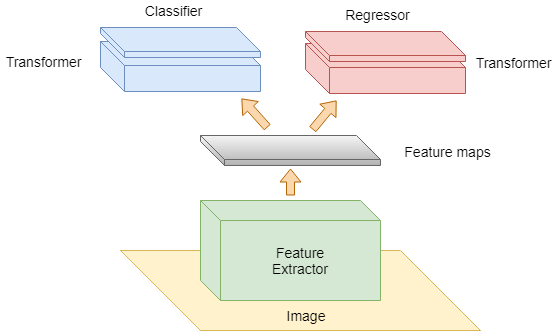}
    \caption{The LS-Net is a feed-forward, fully convolutional neural network and consists of three modules: (i) a fully convolutional feature extractor, (ii) a classifier, and (iii) a line segment regressor.}
    \label{fig:network}
\end{figure}

Inspired by the success of single-shot object detectors such as SSD \cite{liu2016ssd} and YOLO \cite{redmon2016you} in terms of speed and accuracy, we propose a single-shot line segment detector, named LS-Net. The LS-Net is based on a feed-forward, fully convolutional neural network and consists of three modules: (i) a fully convolutional feature extractor, (ii) a classifier, and (iii) a line segment regressor connected as shown in Fig. \ref{fig:network}.

\begin{figure}[h!]
    \centering
    \includegraphics[width=0.40\textwidth]{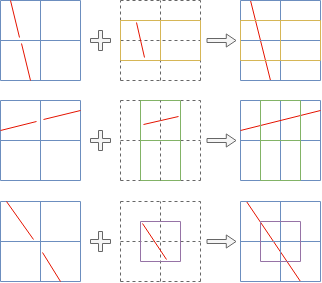}
    \caption{An illustration of the four-grid approach. LS-Net with the traditional one grid approach (the first column) ignores short line segments at cell corners and create gaps at cell borders in the detected lines. LS-Net with the four-grid approach (the third column) utilizes three additional grids (the second column) to detect line segments ignored by the main grid and close gaps at horizontal and vertical borders, which significantly eliminate the discontinuities in the detected lines.}
    \label{fig:4grids_explain}
\end{figure}

\begin{figure*}[t]
    \centering
    \includegraphics[width=0.6\textwidth]{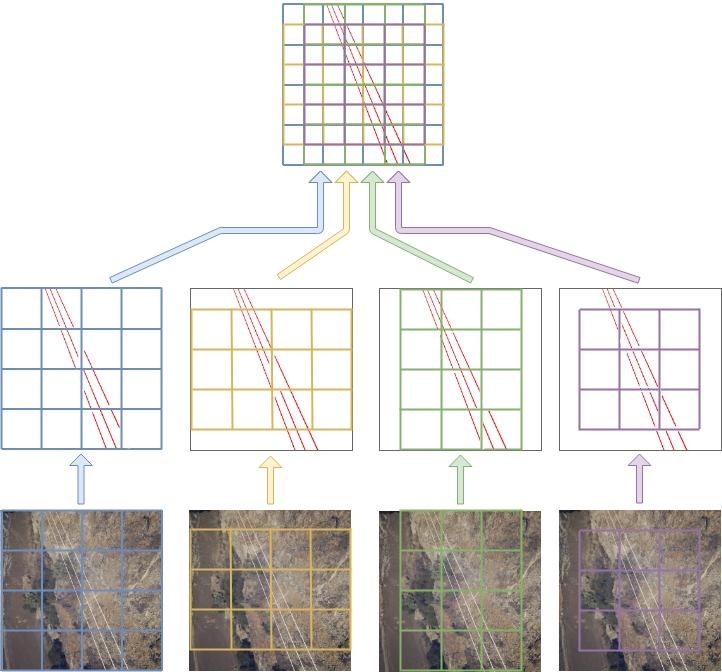}
    \caption{An illustration of the four overlapping grids approach. LS-Net with one grid (the leftmost branch) ignores short line segments at cell corners and create gaps at cell borders in the detected lines. LS-Net with four overlapping grids approach  utilizes three additional grids to detect line segments ignored by the first grid and close gaps in horizontal and vertical lines, which significantly eliminate the discontinuities in the detected lines.}
    \label{fig:flow}
\end{figure*}

{\color{black}The design of the LS-Net architecture is mainly inspired by state-of-the-art single-shot object detectors such as SSD \cite{liu2016ssd} and YOLO \cite{redmon2016you}. Specifically, the LS-Net divides the input image of size $W\times H \times C$  into a grid, and each grid cell of size $C\times C$ predicts coordinates and a confidence score for the longest line segment in the cell. The confidence score indicates the probability of the cell containing a line segment, and the coordinates are the normalized distances of the two endpoints of the line segment to the local $x$-axis and $y$-axis of the cell.

The traditional one grid approach has been proven to work well for single-shot object detectors such as SSD \cite{liu2016ssd} and YOLO \cite{redmon2016you}; however, it faces two problems when applied to line segment detection: (i) discontinuities and gaps at cell borders, and (ii)  discontinuities and gaps at cell corners. In the one grid approach, due to regression errors, the detected line segments can be shorter than the ground truths. This can result in discontinuities and gaps in the detected lines at borders of adjacent cells that make regression errors. In addition, the one grid approach ignores short line segments, especially at cell corners, due to the lack of features. This can also lead to discontinuities and gaps in the detected lines (see Fig. \ref{fig:4grids_explain}).

To address the two above-mentioned problems, we propose to replace the one grid approach by a four-grid approach. Specifically, the four-grid LS-Net divides the input image into four overlapping grids: a $S_m\times S_m$ grid (main grid), a $S_m\times S_a$ grid (horizontal grid), a $S_a\times S_m$ grid (vertical grid), and a $S_a\times S_a$ grid (center grid), where $S_a = S_m - 1$ (see Fig. \ref{fig:flow}). The main grid, which works exactly the same as the grid used by SSD and YOLO for detecting objects, is employed for detecting line segments in grid cells. The horizontal and vertical grids are utilized for closing the gaps at horizontal and vertical borders, respectively. The central grid is used for detecting short line segments at cell corners that were ignored by the main grid. All the detected line segments from the four grids are combined together to form a line segment map. Since the four-grid LS-Net utilizes three additional grids to detect short line segments ignored by the main grid and close gaps at horizontal and vertical borders, the discontinuities in the detected lines are significantly eliminated (see Fig. \ref{fig:flow}).}

\subsubsection{Fully Convolutional Feature Extractor}
The LS-Net feature extractor is inspired by the VGG-16 network \cite{simonyan2014very}. We truncate the network before the last max-pooling layer and substitute the remaining max-pooling layers by strided convolutional layers with stride 2. Max pooling layers have been used extensively in CNNs for image classification; however, they are not an optimal choice for the proposed LS-Net since they throw away spatial information that is useful for predicting line segment end-points.

With the aim of easing the optimization, enabling the network to converge faster, and eliminating the dependence on batch sizes, we adopt Group Normalization \cite{GroupNorm2018} before activations in every convolutional layer.
\subsubsection{Classifier}
The classifier sub-network takes feature maps extracted by the fully convolutional feature extractor as input and predicts whether each grid cell contains a line segment or not. The sub-network consists of two layers: The first is a $2\times 2$ convolutional layer with stride 1 that works as a \textit{transformer (transformation layer)} and transforms the input feature maps into four sets of feature maps corresponding to the four overlapping grids. The second layer is a $1\times 1$ convolutional layer that predicts a confidence score for each grid cell.
\subsubsection{Line Segment Regressor}
The line segment regressor sub-network takes feature maps extracted by the fully convolutional feature extractor as input and predict coordinates of the longest line segment in each grid cell. The sub-network also consists of two layers: The first layer is similar to the first layer of the classifier sub-network. The second is a $1\times 1$ convolutional layer that is responsible for predicting line segment coordinates. 

\subsubsection{Summary}
{\color{black}With the four-grid approach, the output of the LS-Net is very similar to that of a traditional sliding-window detector of size $C\times C$ with stride $C/2$; however, the LS-Net has to major advantages over the sliding-window approach: The first is that instead of applying a costly forward pass hundreds of times, one for each cell, the LS-Net makes predictions for all cells in a single forward pass, which was made possible thanks to the single-shot detector architecture and the combination of our proposed four-grid approach and our proposed transformation layers.  The second advantage is that the LS-Net, with a large effective receptive field, can take into account contextual information when making predictions. In other words, the LS-Net looks at not only the target cell but also its neighboring cells to make predictions for the cell.}

{\color{black}To evaluate the effectiveness of the proposed LS-Net architecture, we train the LS-Net on input images of size $512\times 512\times 3$ to detect line segments in cells of size $32 \times 32$, i.e., $S_m = 16$; however, the proposed LS-Net architecture can be easily generalized to handle images of any sizes and to detect line segments in cells of any sizes. A detailed configuration of the LS-Net used in our experiments in this paper is shown in Table \ref{tab:configurations}. All convolutional layers in the feature extractor are padded so that they produce an output of the same size as the input. Padding is not applied in convolutional layers in the classifier and the regressor.}
\begin{table}[h!]
\centering
\caption{LS-Net's Configuration. The convolutional layer parameters are denoted as ``conv(receptive field size)-(number of channels)[-S(stride)]''. The default stride is 1.} 
\label{tab:configurations}
\begin{tabular}{|c|c|}
\hline
\multicolumn{2}{|l|}{Input image ($512\times 512\times 3$)} \\ \hline
\multicolumn{2}{|c|}{Conv3-64} \\ \hline
\multicolumn{2}{|c|}{Conv3-64} \\ \hline
\multicolumn{2}{|c|}{Conv3-64-S2} \\ \hline
\multicolumn{2}{|c|}{Conv3-128} \\ \hline
\multicolumn{2}{|c|}{Conv3-128} \\ \hline
\multicolumn{2}{|c|}{Conv3-128-S2} \\ \hline
\multicolumn{2}{|c|}{Conv3-256} \\ \hline
\multicolumn{2}{|c|}{Conv3-256} \\ \hline
\multicolumn{2}{|c|}{Conv3-256-S2} \\ \hline
\multicolumn{2}{|c|}{Conv3-512} \\ \hline
\multicolumn{2}{|c|}{Conv3-512} \\ \hline
\multicolumn{2}{|c|}{Conv3-512-S2} \\ \hline
Conv2-512 & Conv2-512 \\ \hline
Conv1-2 & Conv1-4 \\ \hline
\end{tabular}
\end{table}

\subsection{LS-Net Multi-task Loss}
The LS-Net has two sibling output layers. The first sibling layer outputs a discrete probability distribution, $p_t^i = (p^i, 1 - p^i)$, for each grid cell, indexed by $i$, over two classes: \textit{cell with line segments} and \textit{cell without line segments}. The probability distribution $p_t^i$ is computed by a softmax over the two outputs of a $1\times 1$ convolution layer at the $i^{th}$ cell. The second sibling layer outputs coordinates of the two end-points of the longest line segment, $e^i = (e^i_{x1}, e^i_{y1}, e^i_{x2}, e^i_{y2})$, for each grid cell, indexed by $i$. 

Each training cell is labeled with a ground-truth class label $y^i \in \{\pm 1\}$  and a ground-truth end-point regression target $t^i = (t^i_{x1}, t^i_{y1}, t^i_{x2}, t^i_{y2})$. We use a weighted multi-task loss function, $L$, to jointly train for cell classification and line segment end-points regression:
\begin{equation}
\label{equation:multitask_loss}
    L(p_t, y, e, t) = L_{cls}(p_t, y) + \lambda [y=1]L_{reg}(e, t),
\end{equation}
where the Iverson bracket indicator function $[y = 1]$ evaluates to 1 when $y = 1$ and 0 otherwise.

The first task loss, $L_{cls}$, is a Focal loss \cite{lin2017focal} defined as follows:
\begin{equation}
 L_{cls}(p_t, y) =  -\alpha_t (1-p_t)^\gamma \log{(p_t)},
\end{equation}
where $\gamma \geq 0$  is a tunable focusing
parameter,  $\alpha_t \in [0, 1]$ is a weighting factor defined as follows:
 \begin{equation}
 \alpha_t =  
  \begin{cases} 
   \alpha & \text{if } y = 1 \\
   1 - \alpha & \text{otherwise }
  \end{cases},
\end{equation}   
and $p_t \in [0, 1]$ is the model's estimated probability defined as follows:
 \begin{equation}
 p_t =  
  \begin{cases} 
   p  & \text{if } y = 1 \\
   1 - p & \text{otherwise }
  \end{cases}.
\end{equation}  
Since the number of cells without line segments is much larger than the number of cells with line segments, the Focal loss is employed instead of a standard Cross-Entropy loss \cite{Goodfellow-et-al-2016} to address the class imbalance during training.

The second task loss, $L_{reg}$, is a Wing loss \cite{feng2018wing} defined as follows: 
\begin{equation}
    L_{reg}(e,t) = 
      \begin{cases} 
      w\ln{(1 +d/\epsilon)} & \text{if } d < w \\
       d - C & \text{otherwise }
  \end{cases},
\end{equation}
where $w$ is a non-negative upper bound that sets the range of the nonlinear part to $(-w, w)$, $\epsilon$ is a constant that limits the curvature of the nonlinear
region, $C = w - w\ln{(1 + w/\epsilon)}$ is a constant that smoothly links the piecewise-defined linear and nonlinear parts, {\color{black}and $d$ is our proposed error function, which computes the minimum absolute difference between the predicted end-points $e = (e_{x1}, e_{y1}, e_{x2}, e_{y2})$ and the target end-points  $t = (t_{x1}, t_{y1}, t_{x2}, t_{y2})$ defined as follows}:
\begin{equation}
    d(e,t) = \min(\sum(\left |t-e \right |), \sum(\left |t-swap(e) \right |)),
\end{equation}
where $swap(e) = (e_{x2}, e_{y2}, e_{x1}, e_{y1})$ is a function that swaps the order of the two end-points. 

{\color{black}The error function $d$ is employed to allow the LS-Net to predicts the two end-points of a line segment regardless of the order, and} the Wing loss is utilized instead of standard $L_2$ \cite{girshick2014rich} or smooth $L_1$ \cite{girshick2015fast} losses  to restore the balance between the influence of errors of different sizes and to allow the model to regress the line segment end-points more accurately.
\subsection{Training and Testing}
The LS-Net can be trained end-to-end by backpropagation and Stochastic Gradient Descent (SGD) \cite{lecun1989backpropagation}. We implement the LS-Net using the Tensorflow framework \cite{tensorflow}. We train the LS-Net from scratch using the Adam optimizer \cite{kingma2014adam} with initial learning rate 0.0001, 0.9 momentum1, 0.999 momentum2, and batch size 8 (due to memory limitation) on a GeForce GTX 1080 Ti GPU. We use early stopping to prevent the network from overfitting. Our network converges after ~3.5 epochs, which takes around 48 hours of training time. 

Before training, we augment our dataset by generating five random crops and their flipped versions from each image; we further augment the dataset by replacing the background from each image with five randomly selected backgrounds from our background image dataset.  

During training, we apply data augmentation on-the-fly by adding Gaussian-distributed additive noise, by applying Gaussian blur, by performing a series of color manipulations, and by employing elastic deformations. All the on-the-fly data augmentation techniques are applied with a probability of $\alpha$. We use $\alpha = 0.25$ in our experiments.  

{\color{black}For $512\times 512\times 3$ input, the LS-Net runs at 20.4 Frames Per Second (FPS) on a GeForce GTX 1080 Ti GPU at test time. However, the speed can be further increased by employing a shallower, thinner feature extractor and by decreasing the input size.} 

\section{Experiments}
\label{sec:experiments}
\subsection{Comparisons with the State-of-the-Art Results}
\label{sec:comparisionswithsota}

As presented in Section \ref{sec:commonapproachestopld}, there are very few relevant DL-based approaches for power line detection. In addition, two approaches among the four reviewed ones apply deep learning for patch classification only, while the line detection step is still addressed by a traditional line detection or line segment detection algorithm such as the Progressive Probabilistic Hough Transform (PPHT) \cite{li2018power} or the Line Segment Detector (LSD) \cite{gubbi2017new}. This typically results in low analysis speed and a need for post-processing to distinguish between power lines and spurious lines. Since our goal is to facilitate real-time power line detection and avoidance in low-altitude  UAV flights with deep learning, in this section, we compare our proposed LS-Net only to state-of-the-art DL-based approaches for power line detection that offer high analysis speed and require minimal effort in post-processing.

First, we compare our proposed LS-Net with the weakly supervised learning with CNNs (WSL-CNN) approach proposed in \cite{lee2017weakly} on the publicly available Ground Truth of Power line dataset (Infrared-IR and Visible Light-VL) \cite{yetgin2017ground}, which is one of the most widely used power line datasets. The LS-Net and the WSL-CNN approaches share a similar objective that is to localize power lines by using cheaper ground-truth data (GTD) than pixel-level GTD (e.g., image-level class information, line end-points information). For a fair comparison, we convert line segment maps generated by the LS-Net to pixel-level segmentation maps using a similar procedure as applied in \cite{lee2017weakly}. First, the pixel-level segmentation maps, $S$, are generated as follows:
\begin{equation}
   conf(x,y) = max(\{conf(LS_i) \mid (x,y) \in LS_i \}),
\end{equation}
\begin{equation}
    S(x,y) = 
      \begin{cases} 
      0 & \text{ if } (x,y) \not\in LS_i \text{ } \forall i \in [1, L] \\
      conf(x,y) & \text{ otherwise}
  \end{cases},
\end{equation}
where $L$ is the number of detected line segments, $LS_i$ is the list of all pixels belonging to the $i^{th}$ line segment, and $conf(LS_i)$ is a function that returns the confidence score of the $i^{th}$ line segment. {\color{black}Since each line segment predicted by the LS-Net is represented by a pair of two end-points, we apply the 8-connected Bresenham algorithm \cite{bresenham1965algorithm} to form a close approximation to a straight line between the two end-points. We vary the width of the straight line, $W_l$, from 1 to 5 and select $W_l = 2$ and $W_l = 3$ since they result in the highest $F_1$ scores (also known as F-scores or F-measures).} We call these models LS-Net-W2  and LS-Net-W3, respectively. Then, the generated segmentation maps are smoothed by convolving with a two-dimensional Gaussian function, as shown in Ep. (\ref{equation:2dgaussian}). Finally, the predicted segmentation maps are binarized by using the Otsu's method \cite{jianzhuang1991automatic, otsu1979threshold}. The test results are shown in Table \ref{tab:lsnetvswslcnn}. 

Then, we implement the Dilated Convolution Networks for Wire Detection (WD-DCNN) proposed in \cite{madaan2017wire} in Tensorflow. In addition, we improve the WD-DCNN approach by adopting Group Normalization \cite{GroupNorm2018} to accelerate the training of the networks and Focal loss \cite{lin2017focal} for restoring the balance between the influence of errors of different sizes in multiple points regression. We create three improved models. In the first model, we add a group normalization layer after each convolutional layer in the WD-DCNN model (WD-DCNN-GN). We replace the class-balanced Cross-Entropy loss function \cite{yu2015multi}, adopted by the WD-DCNN model, by the Focal loss to train the second model (WD-DCNN-FL). Finally, we combine both Group Normalization and Focal loss to train the third model (WD-DCNN-GNFL). We train the WD-DCNN model and its improved versions on the same training dataset that we use to train our proposed LS-Net. The predicted segmentation maps of the four models are binarized by using the Otsu's method \cite{jianzhuang1991automatic, otsu1979threshold}. We compare against our proposed LS-Net-W2  and LS-Net-W3 models, the WD-DCNN model, and its improved versions (WD-DCNN-GN, WD-DCNN-FL, and WD-DCNN-GNFL) on the Ground Truth of Power line dataset (Infrared-IR and Visible Light-VL) \cite{yetgin2017ground}. The test results are shown in Table \ref{tab:lsnetvswslcnn}.


\begin{table}[h!]
\centering
\caption{{\color{black}Comparisons of the proposed LS-Net-W2  and LS-Net-W3 models and the state-of-the-art DL-based approaches for power line detection including the WSL-CNN, the WD-DCNN, and its improved versions (WD-DCNN-GN, WD-DCNN-FL, WD-DCNN-GNFL) on the Ground Truth of Power line dataset (Infrared-IR and Visible Light-VL). *Results reported by \cite{lee2017weakly}.}}
\label{tab:lsnetvswslcnn}
\begin{tabular}{|l|l|c|c|}
\hline
\multicolumn{1}{|c|}{} & \multicolumn{1}{c|}{ARR} & APR & $F_1$ Score\\ \hline
WSL-CNN \cite{lee2017weakly}* & 0.6256 & - & - \\ \hline
WD-DCNN \cite{madaan2017wire} & 0.7192 & 0.4713  & 0.4835 \\ \hline
WD-DCNN-GN  & 0.8292 & 0.4148  & 0.4882 \\ \hline
WD-DCNN-FL & 0.7514 & 0.4680 & 0.5079 \\ \hline
WD-DCNN-GNFL  & 0.7930 & 0.4690 & 0.5218 \\ \hline
\textbf{LS-Net-W2 ($W_l = 2$)} & 0.7972 & \textbf{0.4874} & \textbf{0.5344} \\ \hline
\textbf{LS-Net-W3 ($W_l = 3$)} & \textbf{0.8525} & 0.4483 & \textbf{0.5256} \\ \hline
\end{tabular}
\end{table}

\begin{figure*}
     \centering
     \begin{subfigure}[b]{0.2455\textwidth}
         \centering
         \includegraphics[width=\textwidth]{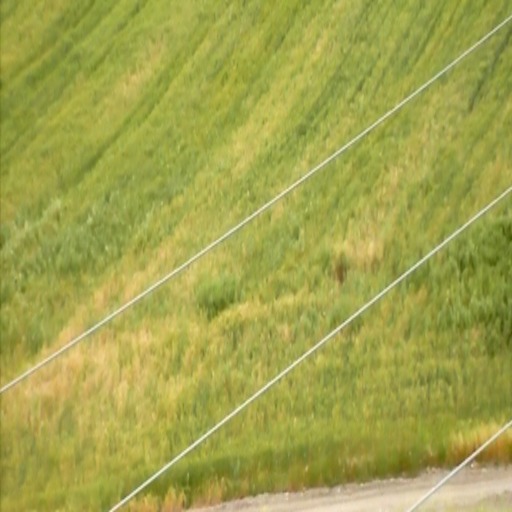}
         \caption{}
     \end{subfigure}
     \hspace*{-0.5em}
     \begin{subfigure}[b]{0.2455\textwidth}
         \centering
         \includegraphics[width=\textwidth]{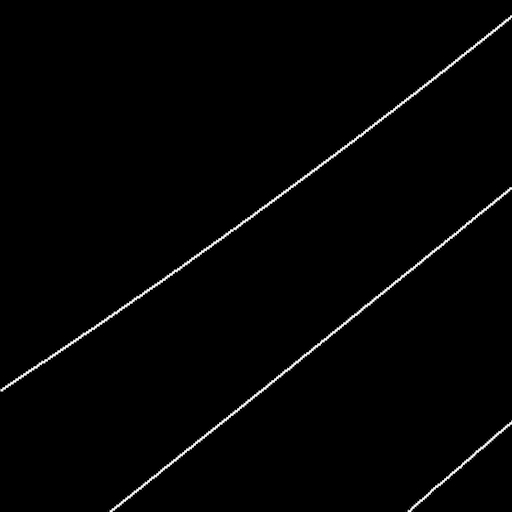}
         \caption{}
     \end{subfigure}
     \hspace*{-0.5em}
     \begin{subfigure}[b]{0.2455\textwidth}
         \centering
         \includegraphics[width=\textwidth]{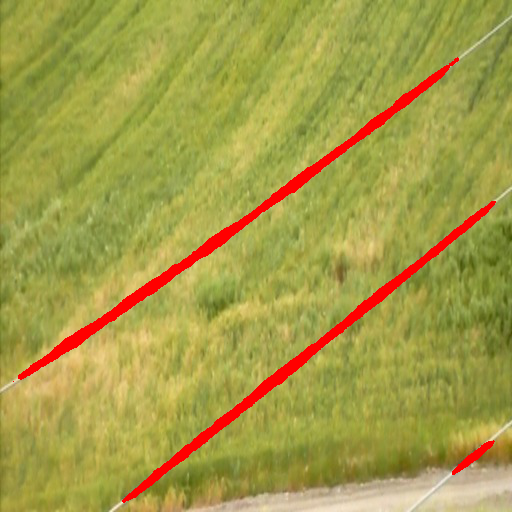}
        \caption{}
     \end{subfigure}
     \hspace*{-0.5em}
     \begin{subfigure}[b]{0.2455\textwidth}
         \centering
         \includegraphics[width=\textwidth]{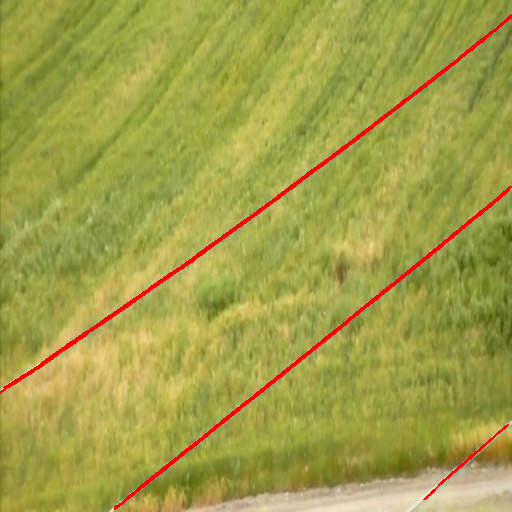}
         \caption{}
     \end{subfigure}
    \vspace*{-0.5em}
    \hspace*{-0.5em}
    
     \begin{subfigure}[b]{0.2455\textwidth}
         \centering
         \includegraphics[width=\textwidth]{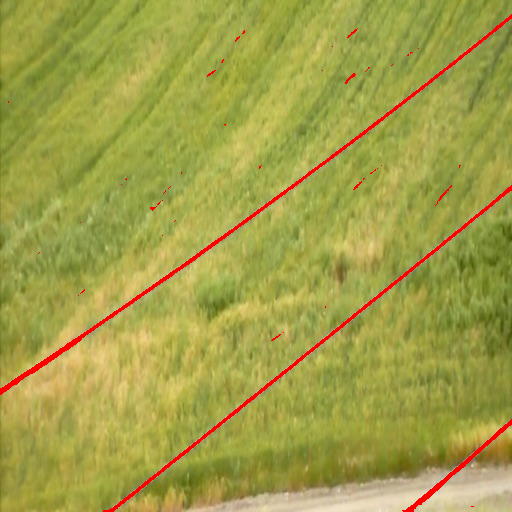}
         \caption{}
     \end{subfigure}
     \hspace*{-0.5em}
     \begin{subfigure}[b]{0.2455\textwidth}
         \centering
         \includegraphics[width=\textwidth]{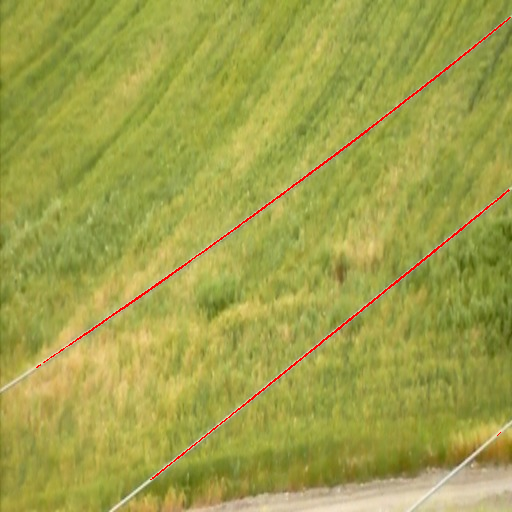}
         \caption{}
     \end{subfigure}
     \hspace*{-0.5em}
     \begin{subfigure}[b]{0.2455\textwidth}
         \centering
         \includegraphics[width=\textwidth]{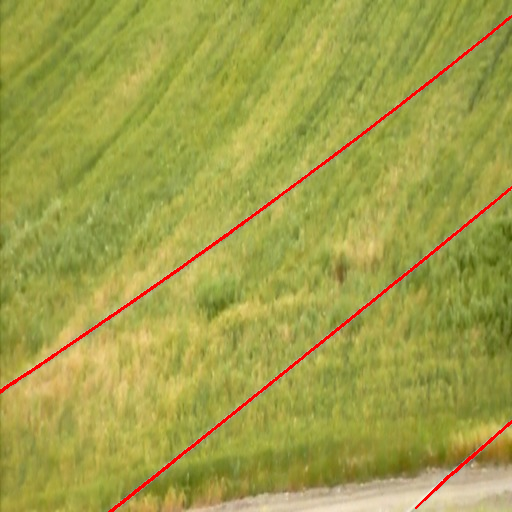}
         \caption{}
     \end{subfigure}
     \hspace*{-0.5em}
     \begin{subfigure}[b]{0.2455\textwidth}
         \centering
         \includegraphics[width=\textwidth]{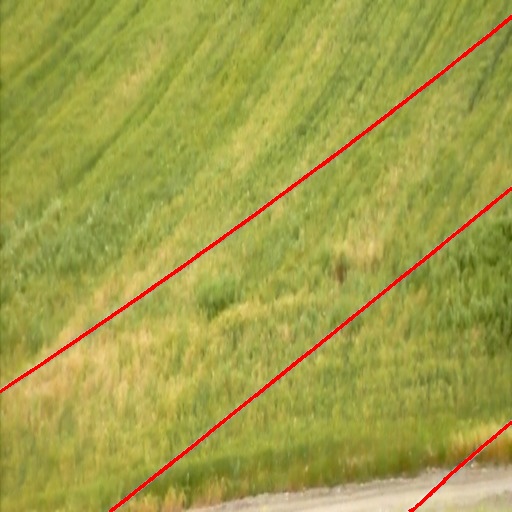}
         \caption{}
     \end{subfigure}
    \caption{{\color{black}Visual comparisons of the LS-Net-W2 model and the state-of-the-art methods. From from left to right, top to bottom are respectively (a) the original image, (b) the ground truth, (c) the WSL-CNN's detection results, (d) the WD-DCNN's detection results, (e) the WD-DCNN-GN's detection results, (f) the WD-DCNN-FL's detection results, (g) the WD-DCNN-GNFL's detection results, and (h) the LS-Net's detection results.}}
    \label{fig:visual_comparisions}        
\end{figure*}

{\color{black}As can be seen from Table \ref{tab:lsnetvswslcnn}, both our proposed LS-Net-W2 and LS-Net-W3 models achieve state-of-the-art performance in terms of $F_1$ score. In addition, the LS-Net-W2 model surpasses all the existing state-of-the-art methods in terms of APR, while the LS-Net-W3 model attains state-of-the-art ARR by considerable margins. Visual comparisons of the LS-Net-W2 model and the state-of-the-art DL-based approaches for power line detection including the WD-DCNN, the WD-DCNN-GN, and its improved versions (WD-DCNN-GN, WD-DCNN-FL, and WD-DCNN-GNFL) are shown in Fig. \ref{fig:visual_comparisions}.}


\subsection{Ablation Study}
To investigate the effectiveness of the proposed LS-Net architecture and the loss function, we conducted several ablation studies using the publicly available Ground Truth of Power line dataset (Infrared-IR and Visible Light-VL) \cite{yetgin2017ground}. We use the approach presented in Section \ref{sec:comparisionswithsota} to convert line segment maps generated by the LS-Net to pixel-level segmentation maps and compare different variants of the LS-Net in terms of APR, ARR, and $F_1$ Score. {\color{black}To increase the interpretability of the comparison results, we apply a simple thresholding method ($t = 0.5$) to binarize segmentation maps instead of the Otsu method and set the width of the line segment $W_l$ to 1 when applying the 8-connected  Bresenham  algorithm}. This could result in lower APR, ARR, and $F_1$ score; however, it is not an issue since improving the performance of the LS-Net is not the primary goal of the ablation studies.

\begin{figure}[h!]
    \centering
    \includegraphics[width=0.48\textwidth]{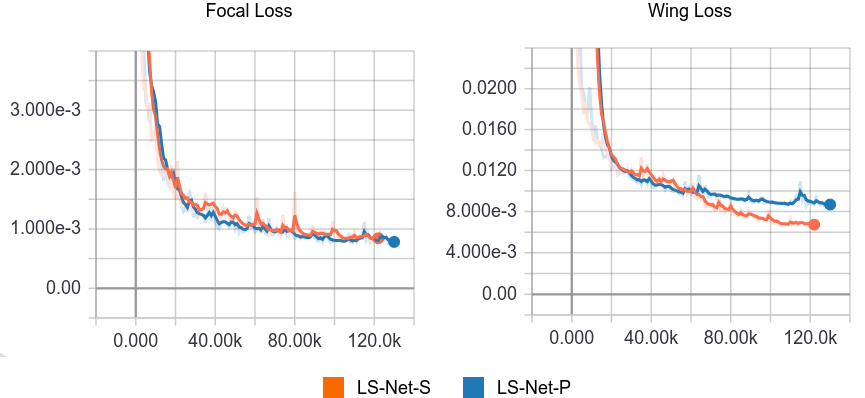}
    \caption{Comparisons between LS-Net-S' and LS-Net-P' test losses. LS-Net-S and LS-Net-P perform similarly on the classification sub-task (Focal loss); however, the LS-Net-S outperforms the LS-Net-P on the line segment regression sub-task (Wing loss)}
    \label{fig:lsnetsvslsnetplosses}
\end{figure}

First, we evaluate the effects of replacing max-pooling layers by strided convolution layers.  To do this, we compare the proposed LS-Net with strided convolutional layers (LS-Net-S) with an LS-Net with max-pooling layers (LS-Net-P), which is constructed by replacing each stride-2 convolutional layer in the LS-Net-S by a stride-1 convolutional layer followed by a max-pooling layer. The comparisons between the LS-Net-S' and the LS-Net-P' performances and losses are shown in Table \ref{tab:lsnetpvslsnets} and Fig. \ref{fig:lsnetsvslsnetplosses}, respectively. As can be seen from Table \ref{tab:lsnetpvslsnets}, the LS-Net-S architecture outperforms the LS-Net-P architecture in terms of APR and $F_1$ Score.

\begin{table}[h!]
\centering
\caption{Comparisons between the LS-Net with strided convolutional layers (LS-Net-S) and the LS-Net with max pooling layers (LS-Net-P).}
\label{tab:lsnetpvslsnets}
\begin{tabular}{|l|l|l|l|}
\hline
Method & APR & ARR & $F_1$ Score \\ \hline
LS-Net-P & 0.7828 & \textbf{0.5378} & 0.5885 \\  \hline
\textbf{LS-Net-S} & \textbf{0.8004} & 0.5368 & \textbf{0.5940} \\ \hline
\end{tabular}
\end{table}
We observe that both LS-Net-S and LS-Net-P perform similarly on the classification sub-task; however, the LS-Net-S outperforms the LS-Net-P on the line segment regression sub-task (see Fig. \ref{fig:lsnetsvslsnetplosses}). This indicates that strided convolution is a more suitable choice for our proposed LS-Net architecture than the standard max pooling.

\begin{figure*}[t]
    \centering
    \includegraphics[width=1.0\textwidth]{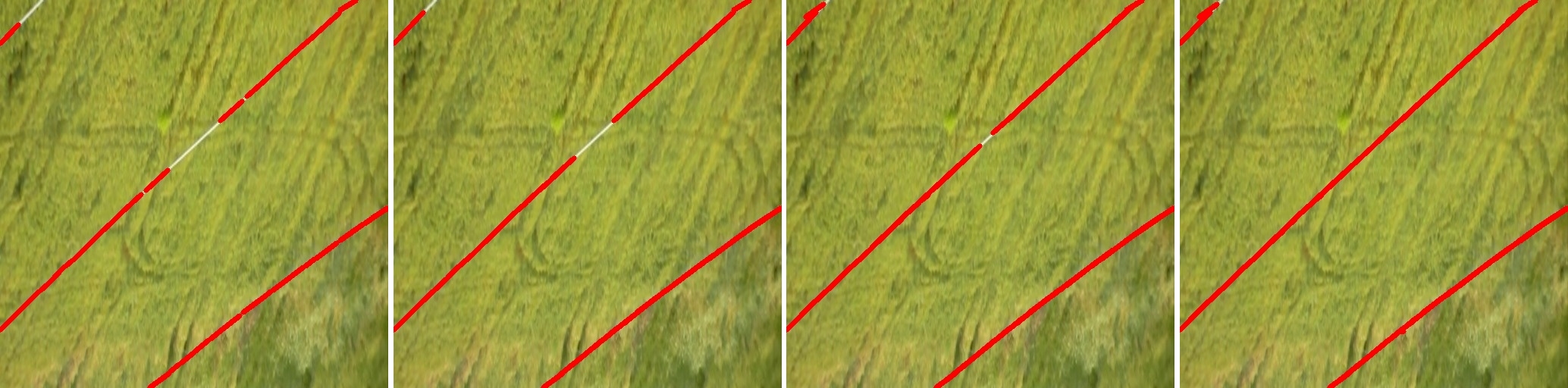}
    \caption{{\color{black}Test results of the LS-Net with (from left to right) one, two, three, and four grids, respectively (the width of the line segments is increased to 5 pixels for better visualizations). The one-grid LS-Net approach (the leftmost image) ignores short line segments at cell corners and leaves gaps at cell borders in the detected lines. The four-grid LS-Net approach (the rightmost image) detects line segments ignored by the first grid and close gaps in horizontal and vertical lines, which significantly eliminate the discontinuities in the detected lines.}}
    \label{fig:grid_test}
\end{figure*}

Then, we investigate the impact of the four-grid approach. We compare against LS-Net with one, two, three, and four grids, respectively. Table \ref{table:grid_test} shows that as the number of grids increases, APR decreases slightly, but ARR increases dramatically. This results in an increase of $F_1$ score as the number of grids increases. Since LS-Net with more grids makes more predictions then LS-Net with fewer grids, their APRs are slightly lower than that of LS-Net with fewer grids. However, as the additional grids detect short line segments ignored by the main grid at cell corners and close gaps at horizontal and vertical borders, the ARR of LS-Net with more girds is significantly higher than that of LS-Net with fewer grids. As can be seen from Table \ref{table:grid_test} and Fig. \ref{fig:grid_test}, the LS-Net with four grids outperforms LS-Net with one, two, and three grids in terms of ARR and $F_1$ score and significantly eliminates the discontinuities in the detected lines. This suggests that our proposed four-grid approach is more suited for our proposed LS-Net architecture.  

\begin{table}[h!]
\centering
\caption{Performance of LS-Net with the one, two, three, and four grids respectively. The methods are denoted as ```LS-Net-$\langle$number of grids$\rangle$-$\langle$grids$\rangle$''. M, H, V, C represent main, horizontal, vertical, and central grids respectively.}
\label{table:grid_test}
\begin{tabular}{|l|l|l|l|}
\hline
Method                 & APR             & ARR             & F1 Score        \\ \hline
\multicolumn{4}{|c|}{1 Grid}                                                 \\ \hline
LS-Net-1-M             & \textbf{0.8312} & 0.3791          & 0.4847          \\ \hline
\multicolumn{4}{|c|}{2 Grids}                                                \\ \hline
LS-Net-2-MH            & 0.8174          & 0.4717          & 0.5540          \\ \hline
LS-Net-2-MV            & 0.8173          & 0.4703          & 0.5533          \\ \hline
LS-Net-2-MC            & 0.8165          & 0.4776          & 0.5574          \\ \hline
\multicolumn{4}{|c|}{3 Grids}                                                \\ \hline
LS-Net-3-MHC           & 0.8080          & 0.5121          & 0.5792          \\ \hline
LS-Net-3-MVC           & 0.8080          & 0.5117          & 0.5791          \\ \hline
LS-Net-3-MVH           & 0.8064          & 0.5165          & 0.5826          \\ \hline
\multicolumn{4}{|c|}{4 Grids}                                                \\ \hline
\textbf{LS-Net-4-MHVC} & 0.8004          & \textbf{0.5368} & \textbf{0.5940} \\ \hline
\end{tabular}
\end{table}

Next, we show the effects of the Wing loss on line segment regression performance. We compare against LS-Net trained with Wing loss (LS-Net-W) and its variants: LS-Net trained with L2 loss (LS-Net-2), L1 loss (LS-Net-1), and Smooth L1 loss (LS-Net-S) \cite{girshick2015fast}, respectively. Table \ref{table:regression_loss} shows that LS-Net trained with Wing loss outperforms its variants in terms of APR; however, it performs worse in terms of ARR.  Wing loss biases the optimizer towards minimizing small regression errors at the end of the training by increasing the gradient, given by $1/x$, as the errors approach zero error. This results in lower regression errors that lead to a significantly higher APR compared to the LS-Net-1, LS-Net-2, and LS-Net-S. However, this causes the classification errors to increase as we use a fixed weight in the multi-task loss (see Eq. (\ref{equation:multitask_loss})). This leads to a slightly lower ARR compared to the LS-Net-1, LS-Net-2, and LS-Net-S. Since the increase in APR is much more than the decrease in ARR, the $F_1$ score of LS-Net trained with Wing loss is higher than LS-Net trained with the standard regression losses such as L2, L1, and Smooth L1. This indicates that the Wing loss is a more suitable choice for training the line segment regressor in our proposed LS-Net architecture; however, an adaptive weighting approach is needed for balancing the training of the line segment regressor and the cell classifier. We leave this for future work. 

\begin{table}[h!]
\centering
\caption{Performance of the LS-Net trained with  Wing loss (LS-Net-W), L2 loss (LS-Net-2), L1 loss (LS-Net-1), and L1 smooth loss (LS-Net-S).}
\label{table:regression_loss}
\begin{tabular}{|l|l|l|l|}
\hline
Method                        & APR             & ARR             & F1 Score        \\ \hline
LS-Net-2 (L2)                 & 0.7277          & \textbf{0.5789}          & 0.5866          \\ \hline
LS-Net-1 (L1)                 & 0.7032          & 0.5694          & 0.5765          \\ \hline
LS-Net-S (Smooth L1)          & 0.7317          & 0.5495          & 0.5728          \\ \hline
\textbf{LS-Net-W (Wing loss)} & \textbf{0.8004} & 0.5368 & \textbf{0.5940} \\ \hline
\end{tabular}
\end{table}

Finally, we evaluate the effect of the Focal loss on cell classification performance. We compare between LS-Net trained with Focal Loss (LS-Net-FL) and LS-Net trained with standard Cross-Entropy loss (LS-Net-CE). As can be seen from Table \ref{tab:focal_loss_test}, LS-Net trained with Focal loss outperforms LS-Net trained with standard Cross-Entropy loss in terms of APR, ARR, and $F_1$ score. This indicates that the Focal loss is a more suitable choice for training the cell classifier in our proposed LS-Net architecture than the standard Cross-Entropy loss.

\begin{table}[h!]
\centering
\caption{Comparisons between LS-Net trained with Focal Loss (LS-Net-FL) and LS-Net trained with standard Cross-Entropy loss (LS-Net-CE).}
\label{tab:focal_loss_test}
\begin{tabular}{|l|l|l|l|}
\hline
Method                          & APR             & ARR             & F1 Score        \\ \hline
LS-Net-CE (Cross Entropy loss)  & 0.7946          & 0.5353          & 0.5899          \\ \hline
\textbf{LS-Net-FL (Focal Loss)} & \textbf{0.8004} & \textbf{0.5368} & \textbf{0.5940} \\ \hline
\end{tabular}
\end{table}


\section{Discussion}
\label{sec:discussion}
With the ability to detect power line segments in near real-time (20.4 FPS), the LS-Net shows the potential to facilitate real-time power line detection and avoidance in low-altitude UAV flights to ensure flight safety. During UAV flights, power line segment maps produced by the LS-Net can be employed to detect power lines and identify dangerous zones quickly, and these information sources can be used as additional inputs to improve the performance of obstacle avoidance and path recovery algorithms. 

In addition, the LS-Net can be utilized for vision-based UAV navigation and for vision-based inspection of power lines. In automatic autonomous power line inspection, the UAV needs to flight along the power lines to take pictures for offline inspections and performs online inspection to identify faults on the power lines (e.g., corroded and damaged power lines) and surrounding objects, such as vegetation encroachment. When GPS-based navigation is not possible, power line segment maps produced by the LS-Net can be employed to navigate the UAV along the power lines. Besides, the power line segment maps can be used for steering the cameras mounted on the UAV to take higher quality pictures of the power lines to improve the performance and reduce the costs of both online and offline inspections.

Since the LS-Net can be trained end-to-end and performs very well even when trained only on synthetic images, it can potentially be adapted for detecting other linear structures. One example is railway track detection. In recent years, the need for automatic vision-based inspection of railway tracks using UAVs has been increasing since UAVs do not require separate tracks for data acquisition as in traditional inspection methods \cite{singh2017vision}. Similar to power line inspection, the LS-Net can be potentially applied for detecting railway tracks from images taken from UAVs. These detections can be utilized both for navigating the UAVs along the railway tracks and for steering the cameras mounted on the UAVs to take pictures of the railway tracks for offline inspections. Another example is unburied onshore pipeline detection in automatic UAV-based gas leak inspection \cite{barchyn2017uav}. Since the width of gas pipelines is relatively big in images taken from UAVs, the LS-Net can not be applied directly to detect gas pipelines. However, this problem can potentially be addressed by casting the gas pipeline detection as a gas pipeline edge detection problem. The LS-Net can be applied for detecting the edges of gas pipelines. The edge detection results can be used for navigating the UAVs along the pipelines, for steering other sensors such as thermal cameras for detecting gas leaks, and even for sizing the pipelines.

In addition to railway track detection and unburied onshore pipeline detection, the LS-Net can potentially be applied for road detection in low- and mid-altitude aerial imagery which facilitates many applications of UAVs such as traffic monitoring and surveillance, path planning, and inspection \cite{zhou2016detecting}. In UAV images, roads are usually very wide; hence, the edge detection approach as used in unburied onshore pipeline detection can be applied.  Roads in satellite images, on the other hand, are usually very narrow and thus can be modeled as lines or curves; this means that the LS-Net can potentially be applied directly for detecting roads in satellite images. 

\section{Conclusion}
\label{sec:conclusion}
This paper introduces LS-Net, a fast single-shot line segment detector. The LS-Net is by design fully convolutional and consists of three modules: (i) a fully convolutional feature extractor, (ii) a classifier, and (iii) a line segment regressor. The LS-Net can be trained end-to-end by backpropagation and stochastic gradient descent (SGD) via a weighted multi-task loss function. The proposed loss function is a combination of Focal loss for addressing the class imbalance in classification and Wing loss for restoring the balance between the influence of errors of different sizes in multiple points regression.

With a customized version of the VGG-16 network as the backbone, the proposed approach outperforms existing state-of-the-art DL-based power line detection approaches. In addition, the LS-Net can run in near real-time (20.4 FPS), which can facilitate real-time power line detection for obstacle avoidance in low-altitude UAV flights, for vision-based UAV navigation and inspection in automatic autonomous power line inspection. Since the LS-Net can be trained end-to-end and performs very well even when trained only on synthetic images, it can potentially be adapted for detecting other linear structures, such as railway tracks, unburied onshore pipelines, and roads from low- and mid-altitude aerial images.

\section*{Acknowledgment}
The authors would like to thank eSmart Systems and UiT Machine Learning Group for support in the work with this paper. This work was supported by the Research Council of Norway [RCN N\AE RINGSPHD grant no. 263894 (2016-2018)  on Power Grid Image Analysis] and eSmart Systems.

\ifCLASSOPTIONcaptionsoff
  \newpage
\fi



%
%

\bibliographystyle{IEEEtran}
\bibliography{references}{}

\end{document}